# Gender Genetic Algorithm in the Dynamic Optimization Problem

P.A. Golovinski, email: golovinski@bk.ru [1]
S.A. Kolodyazhnyi, email: rector@vgasu.vrn.ru [1]

[1] Voronezh State Technical University

A general approach to optimizing fast processes using a gender genetic algorithm is described. Its difference from the more traditional genetic algorithm it contains division the artificial population into two sexes. Male subpopulations undergo large mutations and more strong selection compared to female individuals from another subset. This separation allows combining the rapid adaptability of the entire population to changes due to the variation of the male subpopulation with fixation of adaptability in the female part. The advantage of the effect of additional individual learning in the form of Boldwin effect in finding optimal solutions is observed in comparison with the usual gender genetic algorithm. As a promising application of the gender genetic algorithm with the Boldwin effect, the dynamics of extinguishing natural fires is pointed.

1. Introduction

The optimization problem is standard in finding the best solution. In a mathematically clear form, it appears as a search for the extremum of the function of many variables. Even with the possibility of an exact formulation, the problem does not have an algorithm for guaranteed finding the absolute extremum. One of the most successful methods for solving such problems is the genetic algorithm (GA) proposed by Holland [1], and its numerous modifications [2]. GA is built by analogy with Darwin's theory of natural selection [3] and includes mechanisms of variability, heredity and selection.

Despite the numerous successes in the practical application of GA, it has a number of drawbacks, the main of which is the premature convergence of the process to a local extremum due to the rapid reduction of genetic diversity in the selection process. Another concomitant of this phenomenon is the inability to switch the adaptation mechanism on new important goals, when they appear. Such dynamic tasks are characteristic for controlling the various types of transportation and the development of emergency situations. All this creates a request for more efficient modifications of the GA, less subject to the restrictions.

In population genetics studying natural organisms significant progress has been made in recent decades in understanding the mechanisms of sex and learning in the adaptation of species [4]. It seems promising to transfer these mechanisms to artificial GA. This paper describes the gender genetic mechanism (GGA) with real parameters and learning of individuals in the application to the problems of dynamic optimization with a built-in learning mechanism [5]. It is of particular interest to us as a tool for making optimal decisions in a rapidly changing environment, primarily in determining the balance of forces and means to extinguish forest fires, where resources for solving the problem are always limited, and the situation is developing very quickly.

## 2. Gender genetic adaptation

In a number of theoretical population models, it is assumed that sexual selection enhances common natural selection [6, 7]. Favorable mutations that occur in different individuals are combined by recombination. An asexual population can fix favorable mutations only one after another, while the sexual population establishes them faster, combining, thanks to sexual fixation, many new useful mutations in accordance with the so-called Fischer-Muller effect. The resulting pressure of sexual selection combined with natural selection is a powerful force in accelerating evolution. The rate of adaptation is increased by sexual mechanisms, since sexual selection provides a faster adaptive response in changing conditions, fixing useful mutations [8]. Based on progress in understanding the rate of adaptation, it has been shown that this rate in sex populations is twice as high as in asexual ones. Thus, both theoretical and experimental results of population genetics indicate the advantages of sexual reproduction, compared with asexual, for the survival of a population in a rapidly changing environment.

## 3. Algorithm with the Baldwin effect

Sexual selection is just one of many factors in population genetics. The GGA is based on the concepts of male energy and female choice from population genetics, and uses two different selection schemes simultaneously within the same algorithm [9]. The GGA brings the gender concept to GA from a biological prototype and has visible advantages over classical GAs, due to its flexibility. Standard GAs have many advantages [10], but most of them use the binary variable coding mechanism, which leads to a significant loss of efficiency when searching for solutions in multidimensional continuous spaces. This limits the use of GA in building decision-making systems for managing complex objects. An obvious problem arises, in particular, when a variable can have only a finite number of discrete values, and the binary representation is inadequate. To avoid these limitations, improved genetic algorithms with real coding have been developed aimed at resolving premature convergence in the form of differential evolution [11,12]. The differential evolution algorithm was first proposed by Storn and Price [13].

Like other evolutionary algorithms, the initialization phase is the first step of the GGA. We use a random algorithm to divide the evolutionary population $P$ into two sexual subsets $M$ (male) and $F$ (female), setting the gender symbol of each individual in accordance with the parameter value $p_g$:

$$\mathbf{x} \in M, \text{ if } p < p_g, \tag{1}$$

$$\mathbf{x} \in F, \text{ if } p \geq p_g,$$

where a random real number in the interval [0, 1], denotes the probability that an individual will belong to the subset $M$. In the practical implementation of the algorithm, we assume below $p_g = 0.5$.

The next operator carries out a crossover. There are many types of crossovers with two parents, but a GA with real coding works directly with real variables [14]. For our algorithm, we took a simple arithmetic crossover, where the new values of the variables in the descendants take the average value of the two selected parents. A weighted sum of vector genes of two selected parents is taken with some positive random parameter $\lambda < 1$ for every gene. Let be $\mathbf{x}_i(t)$ is $i$-chromosome generation $t$:

$$\mathbf{x} = (x_{i1}(t), x_{i2}(t), \ldots, x_{in}(t)), \tag{2}$$

where $n$ is the length of the chromosome equal to the number of variables of the fitness function with real coding. Next, let $M, F$ be a subset of male and female parents with a population $N_m, N_f$. Then, for each pair $i \in M, j \in F$, the offspring is defined as follows:

$$\mathbf{z}_i(t+1) = \mathbf{x}_i(t) + \lambda(\mathbf{x}_i(t) - \mathbf{y}_j(t)), \tag{3}$$

where $\mathbf{x}_i(t)$ and $\mathbf{y}_j(t)$ are two randomly selected chromosomes, and $\lambda$ is the random coefficient taken from the interval (0, 1).

Local mutation means the selection of a particular chromosomal vector for a local mutation. As the mutation operator, we will take the operation when all the coordinates of the chromosome change for the random value from a given small range. Mutated vector $\mathbf{x}_i(t+1)$ is calculated using the relation

$$\mathbf{x}_i(t+1) = \mathbf{x}_i(t) + \mathbf{r}, \tag{4}$$

where a random vector $\mathbf{r}$ called a mutation will be small: $|\mathbf{x}_i(t)| \gg |\mathbf{r}|$. We use a mutation in which a random vector $\mathbf{r}$ obeys the Gaussian probability distribution.

We use proportional fitness selection for males, where the level of fitness is related to the probability of choosing each individual chromosome:

$$p_i = \frac{f_i}{\sum_{j=1}^{N_m} f_j}. \tag{5}$$

Here $f_i$ is the fitness of the $i$-th individual, and $N_m$ is the number of individuals in the male subgroup. The objective function $f(\mathbf{x})$ is set by the parameter $\mathbf{x}$ adopted to maximize. As an alternative for females, we use simple random selection.

To guide adaptation through individual acquired fitness, we introduce the Baldwin effect associated with learning [15, 16]. Learning is achieved using the one-step Newton-Raphson method based on the formula

$$\mathbf{x}' = \mathbf{x} - H^{-1} \nabla f(\mathbf{x}) \qquad (6)$$

with Hessian matrix $H$, which immediately gives exact solution for quadratic functions.

We adopt an adaptive genetic strategy according to which the frequency of mutations in women $p_f$ and men $p_m$ decreases gradually and monotonously during the genetic process. The individual level of mutations in women and men is adaptively regulated by the evolutionary generation in accordance with the formulas

$$p_f = p_{f0} \exp(-a_f t / t_{\max}), \qquad (7)$$

$$p_m = p_{m0} \exp(-a_m t / t_{\max})$$

where $p_{f0}$ and $p_{m0}$ are the initial mutation frequency in the female and male subgroups, and $a_f$ and $a_m$ are the positive constants.

The gender-based genetic algorithm with the Baldwin effect is summarized in pseudocode:

(Initial parameters: *population size N, maximum evolutionary generation tmax, mutation parameters pm0, pf0, am, af.*)

1: **Begin**
2: *t*:=0;
3: Initialize[*P*(0)]; (*Generate initial population P(0) of size N at t=0*)
4: **while**(*t<tmax*) **do**
5: [*P*(*t*), *Gender*]:=GD[ *P*(*t*) ]; (*Determine the gender of individuals in population P(t)*)
6: (*pf, pm*):=Rate[*pf0, pm0, am, af, t, Gender*]; (*Adjust mutation rate of female and male subgroups based on evolutionary generation and gender*)
7: *PM*(*t*):=Mutation[*P*(*t*), *pf, pm, Gender*]; (*Gender-based mutation on P(t) to generate population PM(t)*)
8: *PL*(*t*):=Learn[*P*(t), *Fit*(*t*)]; ( *One step Newton–Raphson correction to generate learned population $P_L(t)$* )
9: *Fit*(*t*):= Fitness[*PL*(*t*)]; (*Evaluate the fitness Fit(t) based on learned population*)
11: *P*(*t*+1):=Selection/Mate[*PM*(*t*), *Fit*(*t*), *Gender*]; (*Elitism reservation, selection and gender-based mating on PM(t) to generate new population P(t+1)* )
12: *t*:= *t*+1;
13: **end while**

14: **End**

The Baldwin effect is completely Darwinian in its mechanism. In contrast, even before Darwin, Lamarck suggested that organisms themselves control evolution, and the useful physical characteristics of the obtained phenotype can be transferred back to the organism's genotype. Unrealistic in biology, the evolution of Lamarck becomes a powerful concept in GA, because feedback from phenotype to genotype is a simple computer procedure. The influence of Lamarckism, as well as the Baldwin effect, consists in a noticeable increase in the overall effectiveness of the genetic algorithm. In the Lamarckian version of the algorithm, selection and pairing are carried out with the participation of a learned population.

## 4. Results of computer simulation

We will give an example demonstrating the advantage of BGGA. Our aim is to compare four different time-dependent GA models using the perturbed Rastrigin fitness function [17]:

$$f_R(x, y) = -[20 + x^2 + y^2 - 10(\cos 2\pi x + \cos 2\pi y)] \tag{8}$$

The maximum point of the function has coordinates $(x_{max}, y_{max}) = (0,0)$ and value

$$f_R(x_{max}, y_{max}) = 0. \tag{9}$$

The time-dependent radial perturbation function is taken as

$$g(x, y, t) = A_0(-\lambda t / t_{max}) \exp\left(-\frac{(x - a_x)^2 + (y - a_y)^2}{2\sigma^2}\right), \tag{10}$$

where $\lambda$ is the scale factor of the rate of change in the amplitude of the perturbation over time, $\sigma$ is the parameter of the width of the function, $a_x$ and $a_y$ are the coordinates of the maximum position.

A comparison of numerical experiments with different algorithms for a static objective function shows that the most effective way to adapt is a gender-genetic algorithm with learning individuals in a population. A comparison of Lamarck's computer algorithm and Baldwin's algorithm did not reveal the clear superiority of one of them.

The effectiveness of the GGA depends on the choice of adaptation parameters. Acceleration of the convergence of the process is achieved by selecting the optimal initial value and rate of reduction of mutations. We used GGA optimization of adaptation parameters, i.e. provided meta-learning using a static fitness function, so that the obtained parameters $p_{f0} = 0.37$, $p_{m0} = 0.36$, $a_f = 4.55$, $a_m = 3.57$ are used in a further dynamic problem. To document the higher flexibility of the BGGA in terms of selective pressure, several test runs were performed using the population size of the genetic algorithm $N = 100$ and the maximum evolutionary generation 15. The number of experiments for averaging over the ensemble of realizations is taken equal 500.

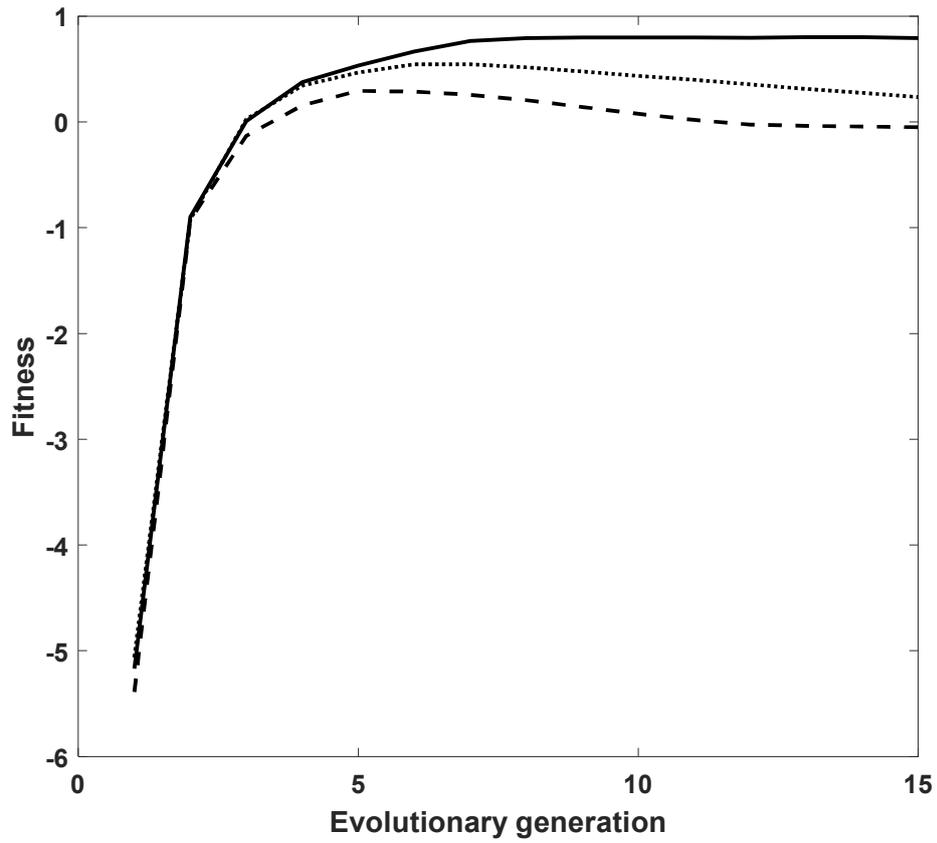

**Figure.** Objective function depending on the generation number.

We evaluate the ability of the BGGA to trace the function maximum in the face of environmental perturbation [18], i.e. the effectiveness of the algorithm in solving the problem of "chasing two birds with one stone". With this formulation of the problem, the rate of change of the environment is controlled by the parameter $\lambda$. The figure shows the dynamics of adaptation for BGGA as applied to the Rastrigin function and the perturbing function $g(x,y,t)$ from equation (10) for $\sigma^2 = 1/40$. The parameters $\lambda = 0.1, 0.5, 1$ correspond to the curves in the sequence from top to bottom. The solid and dot lines shows the algorithm track the decreasing maximum of the perturbation function in the vicinity of the point (0,1). The dashed line indicates the shift of the chase from the maximum of the perturbation function to the maximum of the Rastrigin function near the point with coordinates (0,0). Bifurcation of chase with switching the pursuit of the target occurs with the parameter $\lambda = 0.8$.

In LGAA, bifurcation is observed at $\lambda = 0.9$, i.e. at more noticeable changes in objective function compared with BGGA. This is a result of the larger depletion of genetic diversity in the LGGA algorithm, and, as a result, the algorithm's response to changes in fitness functions slows down. The same result for the bifurcation point is observed in simple GA. For GGA, the bifurcation point is 0.8, as for BGGA, but with a much lower adaptation rate. A comparison of the GA and GGA with the Baldwin and Lamarck algorithms shows that the Baldwin algorithm is the most efficient in this set. It tracks the objective function in the process of its change, providing timely switching and optimization when a new maximum of the objective function appears. This algorithm is one of the best approaches to finding optimal solutions in complex real-time problems.

**Conclusions**

In a comprehensive set of tests, it was shown that BGGA has significant advantages in terms of adaptability of selection pressure. In the classic GA concept, only the size of a tournament group can influence selection pressure. In contrast, when using BGGA, the selection pressure can be controlled more precisely due to the possibilities of combining different selection concepts. Thus, the user of the BGGA can better adjust the algorithm in accordance with the needs of the optimization problem being solved. In addition, the optimization potential contained in the BGGA can be used to a greater extent, since the interaction between the trends of decreasing diversity and factors of its support can be more precisely regulated.

All the algorithms we used in constructing the BGGA were already previously known, however, the high efficiency and robustness of our version of the algorithm is achieved only when using together real coding, gender based selection, uniform differential crossover, adaptive mutation rate based on meta-learning, the Baldwin effect with one-step learning by the Newton-Raphson method, and elitism reservation. Our results, confirming the effectiveness of the BGGA in solving dynamic optimization problems, make it possible to put forward it as an optimization tool to solve the problem of resource allocation in extinguishing forest fires. To this end, it will be necessary to combine it with the forest fire model we are developing, based on the artificial intelligence technologies.